\pdfoutput=1
\documentclass{article} 
\usepackage{iclr2017_conference,times}
\usepackage{hyperref,microtype}
\usepackage{url}

\usepackage{t1enc}
\usepackage[english]{babel}
\usepackage[utf8]{inputenc}

\usepackage{makeidx}  
\usepackage{times,amsmath,amssymb,bm}
\usepackage[pdftex]{graphicx}
\usepackage{subfigure}
\usepackage{color}
\usepackage[normalem]{ulem}
\usepackage{url}
\usepackage{textcomp}
\usepackage{gensymb}
\usepackage{bm}
\usepackage{booktabs,multirow}
\usepackage[group-separator={,},group-four-digits]{siunitx}

\renewcommand{\epsilon}{\varepsilon}

\title{Cognitive Deep Machine Can Train Itself}

\author{A. L{\H o}rincz, M. Cs\'akv\'ari, \'A. F\'othi, Z. \'A. Milacski,  A. S\'ark\'any \& Z. T{\H o}s\'er 
\\
Department of Software Technology and Methodology\\
Faculty of Informatics, E\"otv\"os Lor\'and University\\
P\'azm\'any P\'eter s\'et\'any 1/C, Budapest, Hungary, H-1117
}

\begin{document}

\maketitle

\begin{abstract}
Machine learning is making substantial progress in diverse applications. The success is mostly due to advances in deep learning. However, deep learning \emph{can make mistakes} and its generalization abilities to new tasks are questionable. We ask when and how one can combine network outputs, when (i) details of the observations are evaluated by learned deep components and (ii) facts and confirmation rules are available in knowledge based systems. We show that in limited contexts the required number of training samples can be low and self-improvement of pre-trained networks in more general context is possible. We argue that the combination of sparse outlier detection with deep components that can support each other diminish the fragility of deep methods, an important requirement for engineering applications. We argue that supervised learning of labels may be fully eliminated under certain conditions: a component based architecture \emph{together} with a knowledge based system can train itself and provide high quality answers. We demonstrate these concepts on the State Farm Distracted Driver Detection benchmark. We argue that the view of the \emph{\citet{AI2030}} may overestimate the requirements on `years of focused research' and `careful, unique construction' for `AI systems'.
\end{abstract}



\section{Introduction}\label{s:intro}

Machine learning is progressing quickly due to deep learning. The key tool for deep learning is crowd-sourcing, i.e., to the exploitation of human intelligence. Success stories demonstrate that superhuman performance can be reached this way \citep{schmidhuber2015deep}. Still, the groundbreaking deep network approach seems limited as `each application requires years of focused research and careful unique construction' \citep{AI2030}. However, if we take a look at human information processing, for example, we learn that it has two basic routes: (i)  holistic recognition \citep{tanaka2011features} and (ii) recognition by components \citep{biederman1987recognition}. These processing methodologies are competing and also complementing each other. Deep learning methods, on the other hand, tend to favor end-to-end learning, which corresponds to holistic recognition and are missing the advantages of the component based approach.

Furthermore, holistic recognition and thus end-to-end learning is fragile. The fragility has been shown in a number of studies., e.g., (i) deep networks can be fooled as described by \citet{nguyen2015deep}, when inputs that differ enormously for a human observer from a given class are assigned the label of that class with extremely high confidence, or as demonstrated in \citet{sharif2016accessorize} showing that (ii)  barely visible watermark-like modifications may change the class label, and that (iii) small additional components can be designed to change the class index to another desired one, making the network prone to attacks. One may admit that the highly sophisticated human visual recognition system is also prone to illusions and can misinterpret visual information. 

\begin{figure}
\centering
\subfigure[]{\includegraphics[height=35mm]{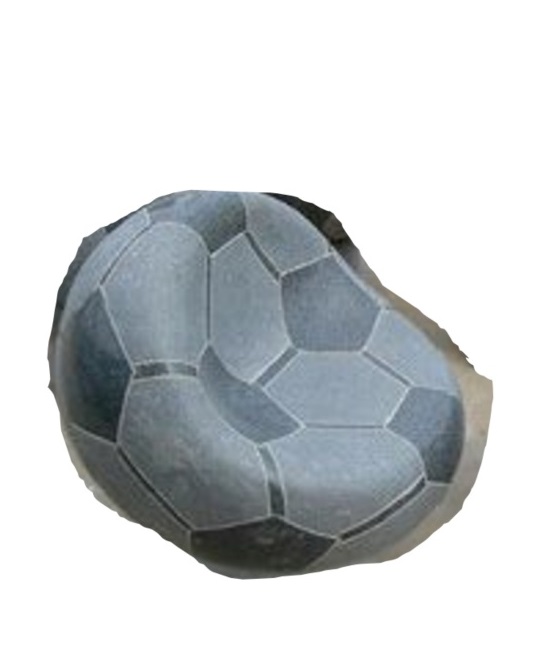}\label{a}}\hspace*{15mm}
\subfigure[]{\includegraphics[height=35mm]{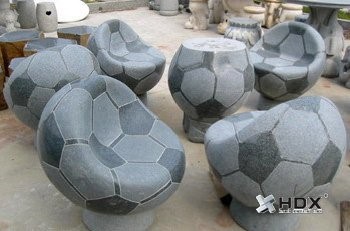}\label{b}}\\
\subfigure[]{\includegraphics[height=30mm]{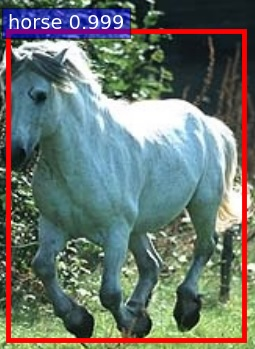}\label{c}}
\subfigure[]{\includegraphics[height=30mm]{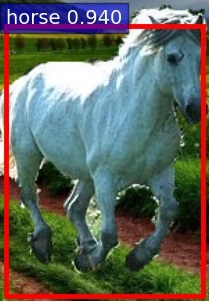}\label{d}}
\subfigure[]{\includegraphics[height=30mm]{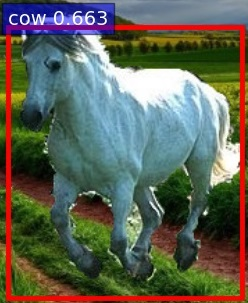}\label{e}}
\subfigure[]{\includegraphics[height=30mm]{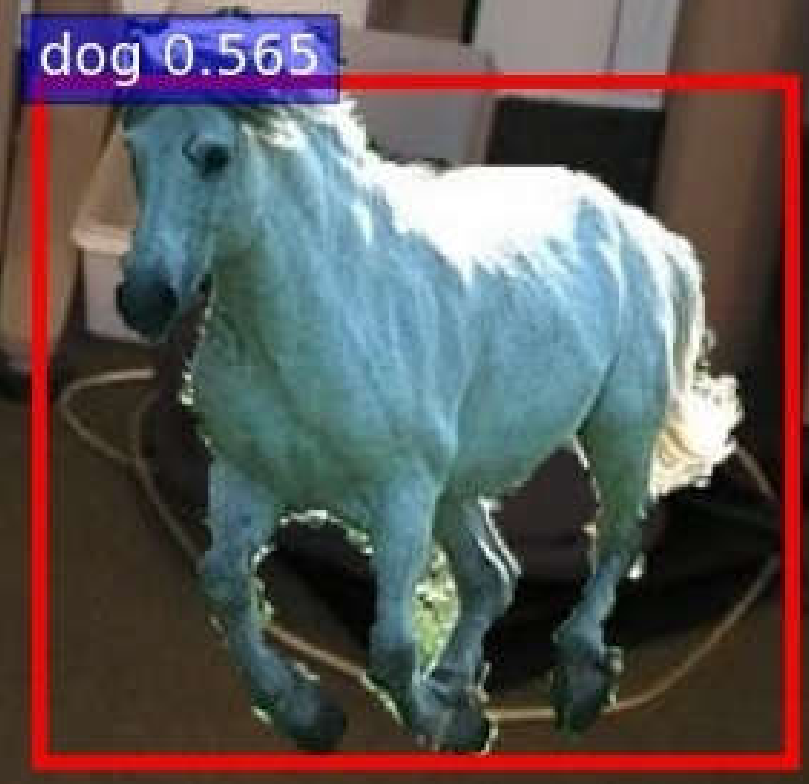}\label{f}}
\subfigure[]{\includegraphics[height=30mm]{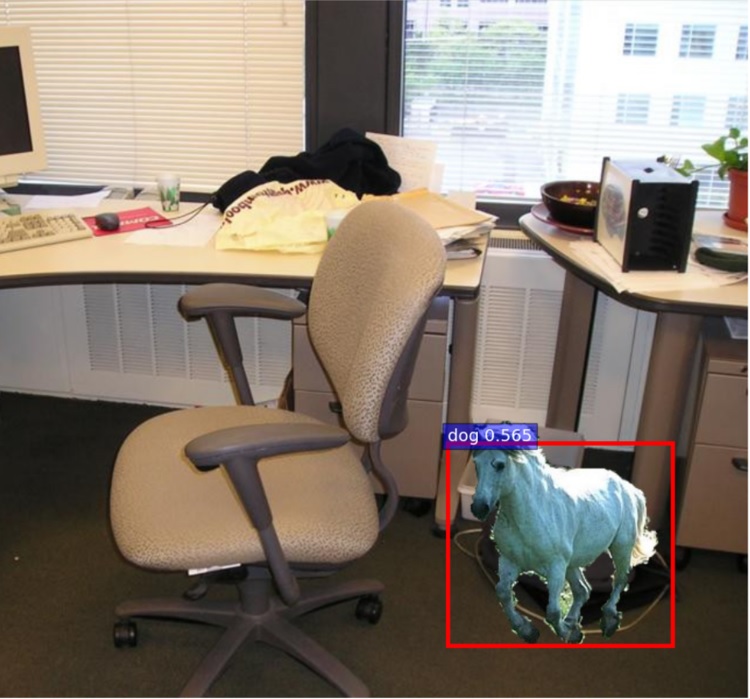}\label{g}}
\caption{Objects and contexts. (a): segmented object, (b): same object in its environment, (c): MS COCO training sample, (d)-(f): segmented horse placed into different environments, (g): same as (f), but zoomed out. Red: bounding box, white characters in blue background: best network proposals with the belonging scores.}
\label{fig:semantics}
\end{figure}

An additional and apparent source of error is the dependency on the context. This is the tool that -- eventually -- we want to exploit for diminishing the problems mentioned above. We illustrate the problem through a few examples. Consider Fig.~\ref{fig:semantics}. Figure~\ref{a} looks like a deflated football, but the interpretation can be changed easily (see Fig.~\ref{b}) by varying the environment, it can also be a chair. Deep learning can make similar `mistakes': we used the Faster R-CNN network that was pre-trained on the Pascal VOC database. The training set contains a horse (Fig.~\ref{c}). We changed the environment of the horse without adding any occlusion. The new backgrounds were chosen from the Visual Genome database. We provide a few samples of our results. For example, the best guess remained `horse' when we put the horse along a road (Fig.~\ref{d}), but it changed to `cow' upon mirroring the horse from left-to-right (Fig.~\ref{e}). Inserting the segmented horse to the bottom of an office desk, `dog' was the best guess of the network (Fig.~\ref{f}). Intriguingly, these network proposals make sense due to the context: typically cows are crossing roads and dogs are in the office. Somehow, somewhere, such knowledge is implicitly embedded into the database. Our aim is to make it explicit.

Why is it so easy to change the output of the network trained on many examples? On the one end, it is good, since context can help, e.g., when words are ambiguous. On the other end, in case if the category is certain then such uncertainty seems worrying. We put forth the idea that holistic recognition can be disambiguated by the recognition of components, such as the mane or the hoof for the case of the horse in our example. Such component based reasoning can easily fix the output of the network. Furthermore,  if time series are available, then the motion pattern is of help since it differs considerably for horses, dogs, and cows. Components have several advantages: (i) they are in smaller spaces, (ii) their environment may be correlated with them, like the example of face being the `environment' of the eyes and the mouth, and (iii) temporal continuity, if present, may improve the precision of the observation process. Nonetheless, cooperation and competition between holistic information processing and component based inference may not overcome all problems as supported by different visual, or auditory illusions and certain combinations, like the McGurk effect \citep{mcgurk1976hearing}, too. Note also that from the general information processing point of view, components are useful; we can acquire additional knowledge by hearsay and can connect lower level component detectors and higher level symbols offering a  solution to the symbol grounding problem emphasized by \cite{harnad2003symbol}. Holistic and component based recognition mechanisms that seem to compete and complement each other point to sophisticated hierarchical construction beyond present day deep neural network architectures.

We demonstrate that traditional knowledge based systems are capable of bridging and training deep neural networks working on different, but correlated components of a larger recognition problem. In general, knowledge based systems may include reasoning tools, differential equations, knowledge about the physics of the world, ontologies, among others. An important ingredient of our approach is the condition that there are \emph{components that assume each other}\footnote{The task of learning of such components has been tackled recently by \cite{lorincz2016estimating}.}. Our procedure has three building blocks. The first one is Robust Principal Component Analysis (RPCA) that filters out apparent components, which are, in fact, outliers. A related problem, Group Fused LASSO is capable of segmenting time series in the presence of non-Gaussian, but sparse deviations offering, e.g., the capability of grouping samples in time. The second part is self-training: the individual deep learning architectures pre-trained in general scenarios can fine-tune each other in a limited context. The third element is the derivation of rule based systems from the episodic labels that are given even if the meaning of the labels is hidden from us. We illustrate our approach on the State Farm Distracted Driver Scenario Kaggle benchmark.

Theoretical background about the procedures we used is given in Sect.~\ref{s:back}. We treat the driver monitoring benchmark in Sect.~\ref{s:results}. We show that the deep components, the outlier detection and the rule based system together that we call Cognitive Deep Machine (CDM) improves performance. The discussion (Sect.~\ref{s:disc}) considers the generality of the results. In this section we argue that development of novel engineering solutions  exploiting AI and deep learning may be \emph{much faster than expected} by the \emph{\citet{AI2030}}. This is due to two reasons: deep networks can be (a) reused, can strengthen and train each other in narrow contexts and (b) connected to high level (classical) rule-based expert systems. In Section~\ref{ss:cognition} we elaborate this concept within our framework. A short summary concludes this technical report (Sect.~\ref{s:sum})

\section{Methods}\label{s:back}

Below, we describe the outlier detection and temporal segmentation schemes (Sect.~\ref{ss:rpca}--\ref{ss:group_lasso}). We discuss Optical Flow and its unsupervised way of finding components (Sect.\ref{ss:OF}). It is followed by the list of pre-trained deep networks that we use and fine-tune through self-supervision (Sect.~\ref{ss:dns}).

\subsection{Robust Principal Component Analysis}\label{ss:rpca}
It is well known that classical Principal Component Analysis (PCA) essentially works with $\|\cdot\|_F$ (Frobenius) norm estimation, and hence it breaks down in presence of additional gross-but-sparse outliers. \cite{candes2011robust} showed that it is possible to augment this architecture with a term that collects and thus separates such components, which they call Robust Principal Component Analysis (RPCA). Accordingly, given $\mathbf{X}\in\mathbb{R}^{D \times T}$, one may define the following convex optimization problem:
\begin{equation}
    \min_{\mathbf{U},\mathbf{S}}{\frac{1}{2}\|\mathbf{X}-\mathbf{U}-\mathbf{S}\|_F^2+\lambda \|\sigma(\mathbf{U})\|_1+\mu \|\mathrm{vec}(\mathbf{S})\|_1},
\end{equation}
i.e., $\mathbf{U}\in\mathbb{R}^{D \times T}$ approximates $\mathbf{X}$ with low-rank (via the first $\ell_1$ regularizer for its singular value vector $\sigma(\mathbf{U})$), while $\mathbf{S}\in\mathbb{R}^{D \times T}$ represents an extra sparse outlier term (due to the second $\ell_1$ regularizer). We utilized the Inexact Augmented Lagrangian Multiplier  \citep{lin2010augmented} solver for this problem implemented in Python\footnote{\url{https://kastnerkyle.github.io/posts/robust-matrix-decomposition/}}. The RPCA method does not require time series, but we utilized it this way.

\subsection{Group Fused LASSO}\label{ss:group_lasso}
Convex multiple change point detection for multivariate time series relies on $\ell_{1,2}$ regularized Frobenius norm estimation as well, where the regularizer acts on the finite difference of the optimization variable, i.e., for input $\mathbf{X}\in\mathbb{R}^{D \times T}$ and weight matrix $\mathbf{W}\in\mathbb{R}^{D \times T}$, solve:
\begin{equation}
    \min_\mathbf{V}{\frac{1}{2}\|\mathbf{W}\circ(\mathbf{X}-\mathbf{V})\|_F^2+\lambda \sum_{t=1}^{T-p}{\|\mathbf{V}\mathbf{Q}_{.,t}\|_2}}
\end{equation}
This problem is often called the Group Fused LASSO \citep{bleakley2011group}. Here, $\circ$ indicates the elementwise product and $\mathbf{Q}\in\mathbb{R}^{T \times T-p}$ is a finite differencing matrix that differentiates $\mathbf{V}\in\mathbb{R}^{D \times T}$ $p$ times, yielding a piecewise polynomial model of degree $p-1$. The $\ell_{1,2}$ regularizer promotes joint vanishing of finite difference components of individual time steps. We implemented the above problem in CVXPY\footnote{\url{http://www.cvxpy.org/en/latest/}} \citep{diamond2016cvxpy}.

\subsection{Optical Flow and component search}\label{ss:OF}

We used Optical Flow \citep{lucas1985optical} implemented in OpenCV\footnote{\url{http://opencv-python-tutroals.readthedocs.io/en/latest/py_tutorials/py_video/py_lucas_kanade/py_lucas_kanade.html}} for searching components across time via estimating bounding box motion between neighbouring frames in video. First, we applied an object detector (Sect.~\ref{ss:dns}) to obtain the bounding box coordinates for each frame, then the boxes were scaled to identical sizes. Next, for the original frames, we estimated the motion of the feature points within the bounding boxes. This provided a similarity measure between consecutive frames. If similarity was above a threshold, we grouped bounding boxes together. We kept doing this for multiple time steps. Finally, we also merged groups that were similar.

The original method of \citep{lucas1985optical} has been improved in many ways, including pyramidal evaluations that start from low resolution and work towards higher ones, Dense Optical Flow methods\footnote{\url{http://docs.opencv.org/3.0-beta/modules/optflow/doc/dense_optflow.html}} and deep architectures, also in combinations, see, e.g., the work of \citet{Fischer15flownet} and the references therein. The 30 years old method serves illustration purposes here.

\subsection{Pre-trained Supervised Deep Networks}\label{ss:dns}

Deep learning is thoroughly reviewed by \citet{schmidhuber2015deep}. Introduction and details of the theory of the different networks can be found in the very recent book from \citet{Goodfellow-et-al-2016-Book}. Consequently, we refer the interested reader to these excellent works and restrict this technical report to the collection of the references of the papers and the related software tools that we applied during the course of our work:
\begin{enumerate}
    \item Faster R-CNN\footnote{Code: \url{https://github.com/rbgirshick/py-faster-rcnn}} \citep{ren2015faster} for object proposals;
    \item Region-based Fully Convolutional Network\footnote{Code: \url{https://github.com/Orpine/py-R-FCN}} \citep{DaiLHS16} for object and hand detection;
    \item Convolutional Pose Machine\footnote{Code: \url{https://github.com/shihenw/convolutional-pose-machines-release}} \citep{wei2016convolutional} for body pose detection (we also injected hand detection for improving the heat maps of this network);
    \item Libfacetracker \citep{toser2016personalization} for face detection.
\end{enumerate}

The networks were pre-trained on the following databases:
\begin{enumerate}
    \item MS COCO\footnote{Database: \url{http://mscoco.org/}} \citep{lin2014microsoft} for object detection;
    \item PASCAL Visual Object Classes\footnote{Database: \url{http://host.robots.ox.ac.uk/pascal/VOC/voc2012/index.html}} \citep{everingham2010pascal} with twenty classes for training Faster R-CNN;
    \item Visual Genome\footnote{Database: \url{https://visualgenome.org/}} \citep{krishna2016visual} for changing backgrounds;
    \item Mittal’s Hand Dataset\footnote{Database: \url{http://www.robots.ox.ac.uk:5000/~vgg/research/hands/index.html}} \citep{mittal2011hand} for hand detection;
    \item VIVA Hand Detection Dataset\footnote{Database: \url{http://cvrr.ucsd.edu/vivachallenge/index.php/hands}} for left and right hand classification.
\end{enumerate}

\subsection{Kaggle State Farm Distracted Driver Detection Dataset}
We performed and evaluated our methodology on the Kaggle State Farm Distracted Driver Detection challenge that was finished a few months ago on August 1, 2016\footnote{Database: \url{https://www.kaggle.com/c/state-farm-distracted-driver-detection}}.
The database contains 26 subjects and 1 video for each. The original benchmark consisted of individual images but was later sorted and assembled into videos by Gilberto Titericz Junior\footnote{\url{https://www.kaggle.com/titericz/state-farm-distracted-driver-detection/just-relax-and-watch-some-cool-movies/code}}. The recordings consist of driving scenarios viewed from the passenger's seat with the camera facing the driver. The world outside the window is hazy and light. The chair, the body of the driver and the dashboard barely move.

\begin{figure}[ht!]
\centering
\subfigure[]{\includegraphics[width=120mm]{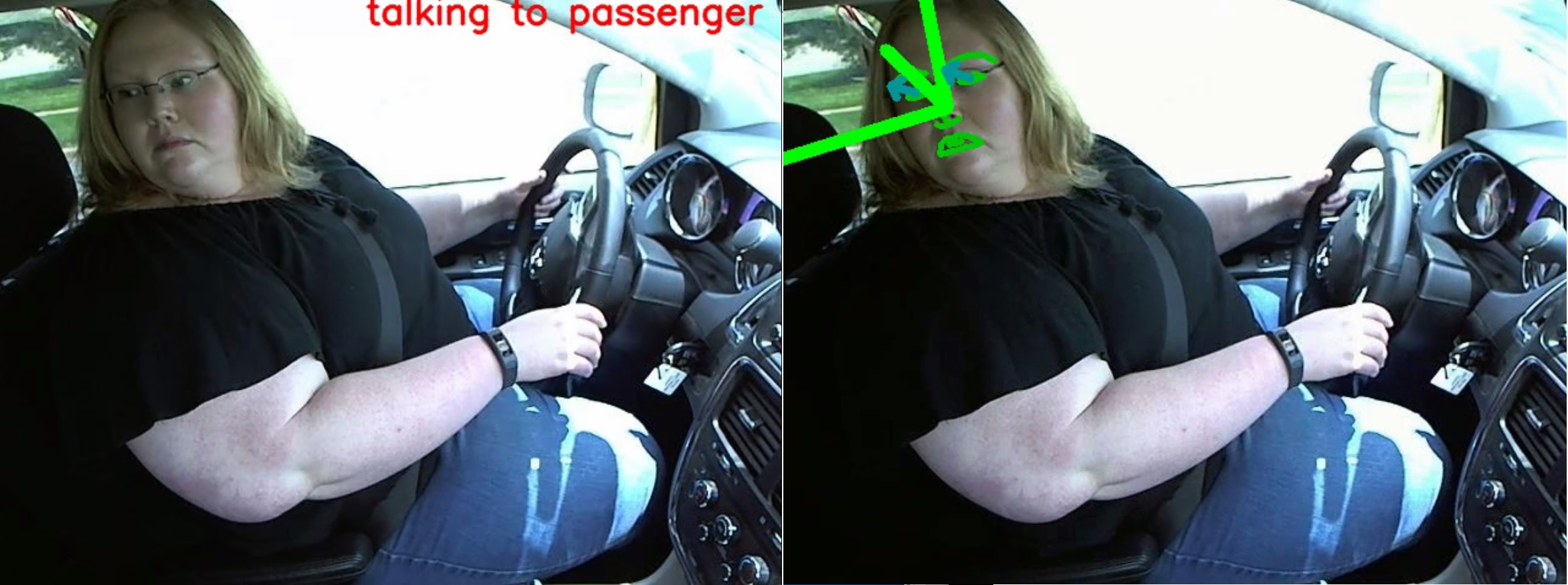}\label{vz}}
\caption{Facial expression and gaze direction estimation in the Kaggle benchmark. The `talking to passenger' category may not be inferred  from a single image and in the absence of acoustic signals (see text). Narrow green lines connect facial markers, broad green lines represent head pose, turquoise arrows starting at the iris show estimated gaze directions.}
\label{fig:talking}
\end{figure}

Labels of the State Farm Distracted Driver Detection challenge are ambiguous and can be misleading. For example, the image of Fig.~\ref{fig:talking} may correspond to safe driving when the driver is looking back to overcome the limits of the mirrors and monitor the blind spots of them directly. On the other hand, there are samples with safe driving labels, when the driver is looking at the passenger and is laughing, she is clearly distracted. A rule based system that takes into account if the driver is talking and the estimation of gaze direction would classify the latter as `talking to passenger'. Decision about the correct label of Fig.~\ref{fig:talking} may require more information about the past and the goals of the driver. For example, she may want to change lanes and may be asking the passenger on the back seat to move so she can see the blind spot of the mirror of the vehicle. There are other examples where labeling is ambiguous, e.g., if the driver is texting and talking  simultaneously.

\section{Experiments and Results}\label{s:results}

\subsection{Pixel based Robust Principal Component Analysis}\label{ss:pixel_anomaly}
\begin{figure}[ht!]
\centering
\subfigure[]{\includegraphics[height=17mm]{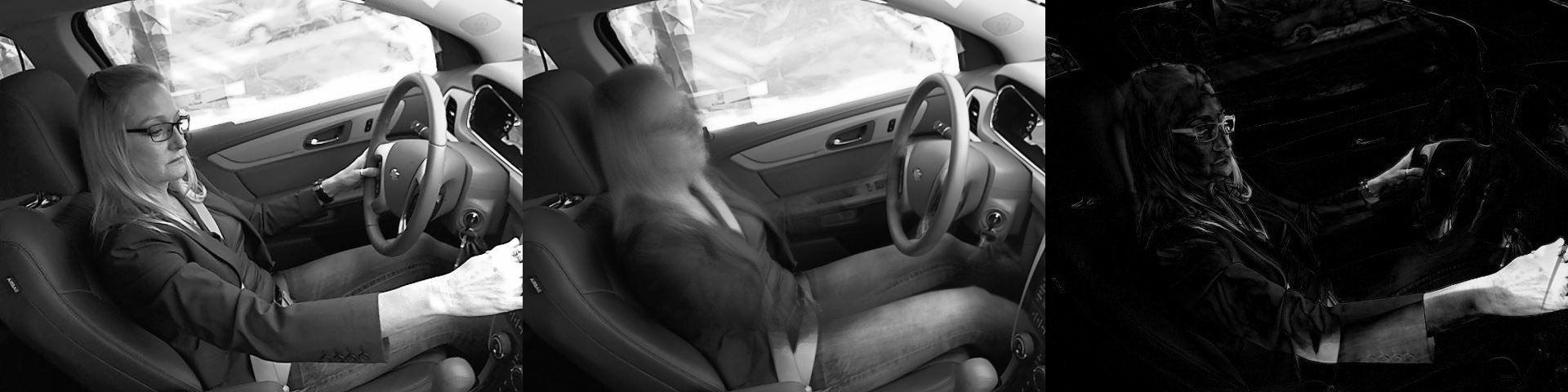}\label{ra}}
\subfigure[]{\includegraphics[height=17mm]{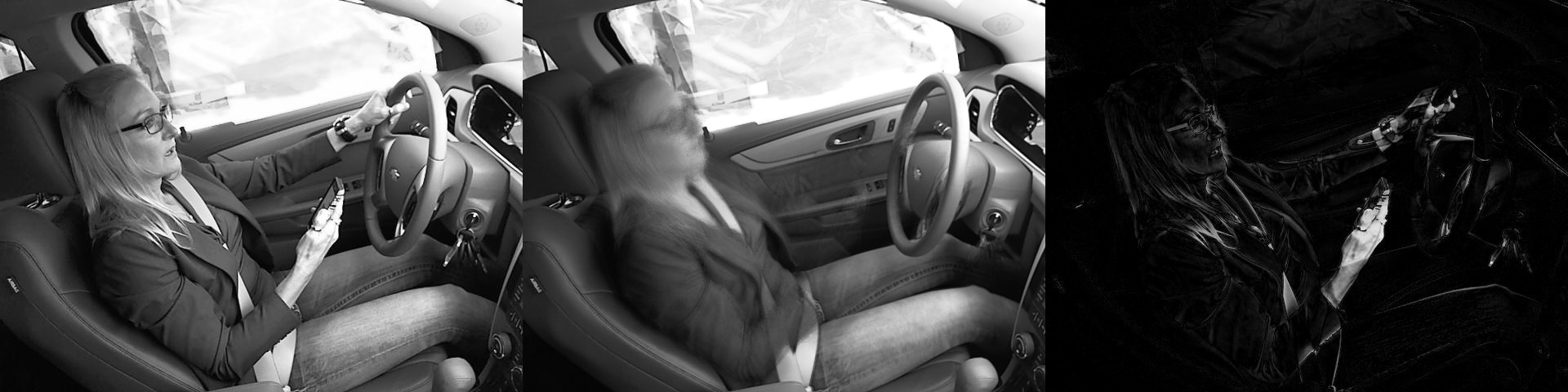}\label{rb}}\\
\subfigure[]{\includegraphics[height=17mm]{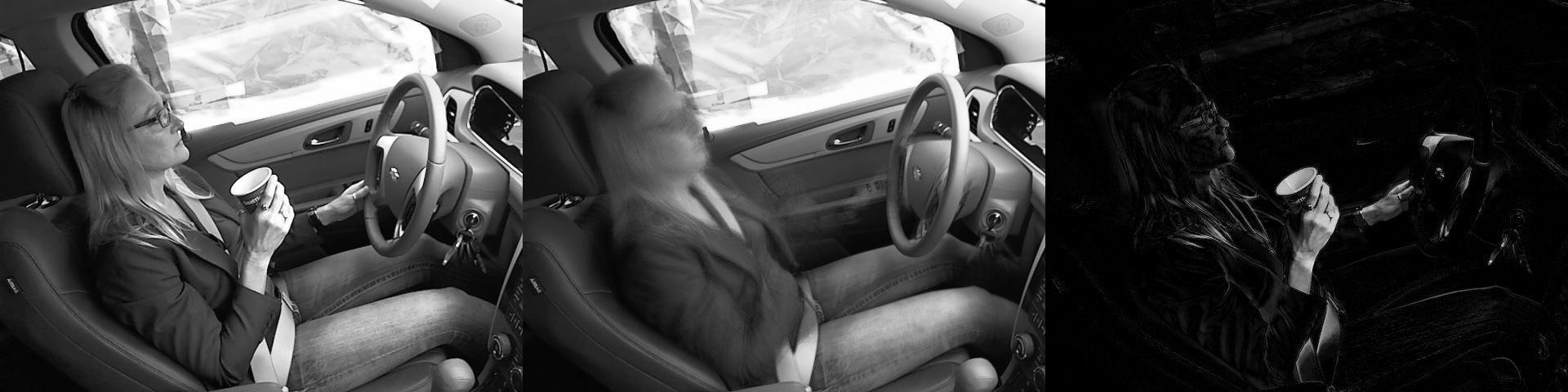}\label{rc}}
\subfigure[]{\includegraphics[height=17mm]{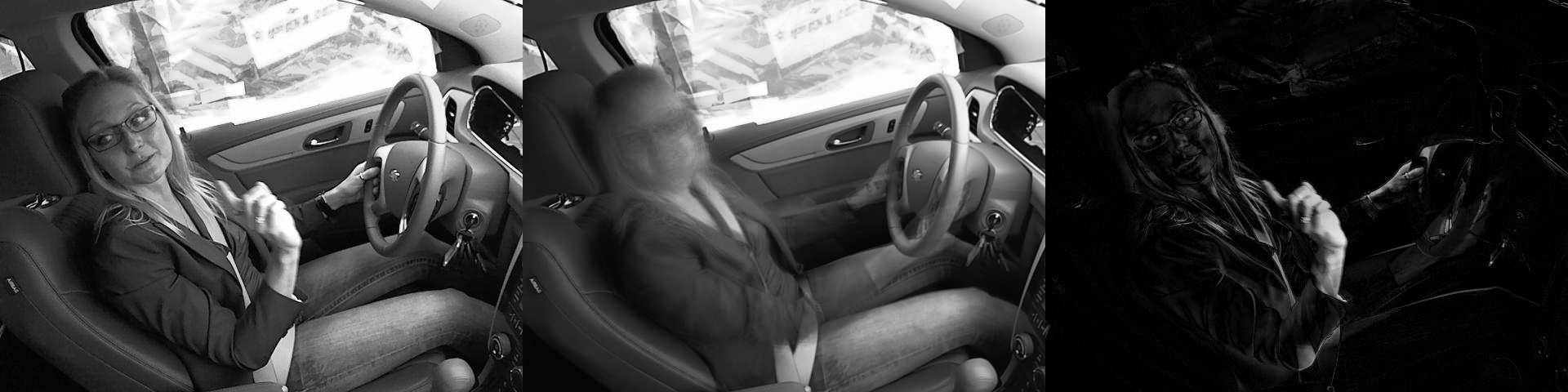}\label{rd}}
\caption{The effect of Robust Principal Component Analysis (RPCA). In each subfigure, the left, middle and right images represent the original, the typical part and the outlier components, respectively. The latter for the right hand are as follows. (a): hand is on the dashboard, (b): with mobile, (c): hand with paper cup, (d): hand in calling attention position. The algorithm may collect atypical body or head poses or objects in the outlier component, and fill them in within the low rank subspace using superposition.}
\label{fig:RPCA1}
\end{figure}

Below, we show examples from the RPCA analysis (Sect.~\ref{ss:rpca}) applied for each grayscale video independently with default parameters. Figure~\ref{fig:RPCA1} has four sets of three subfigures. The left, middle and right subfigures represent the original image, its low rank and its outlier components, respectively.
A large portion of the images is stationary; these parts form the low dimensional subspace of the RPCA analysis. The outliers mostly correspond to short time intervals within driving episodes that are called, for example, `operating the radio' (\ref{ra}), `texting with the right hand' (\ref{rb}), `drinking' (\ref{rc}), `talking to the passenger' (\ref{rd}) and a few more. Note that the outlier parts as well as the low dimensional subspace parts differ: they are projections onto the PCA subspace and these projections depend on the input.
\subsection{Episodic segmentation with Group Fused LASSO}\label{ss:PM_episodic}
Similar to outliers being correlated with ground truth labels in Sect.~\ref{ss:pixel_anomaly}, we note that driver body poses are also associated. Such information can be determined by means of a pose detector such as the Convolutional Pose Machine (Sect.~\ref{ss:dns}).

\begin{figure}[ht!]
\centering
\subfigure[]{\includegraphics[width=60mm]{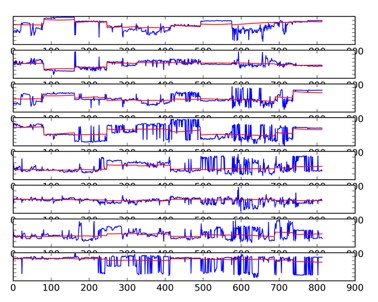}\label{s1}}
\subfigure[]{\includegraphics[width=120mm]{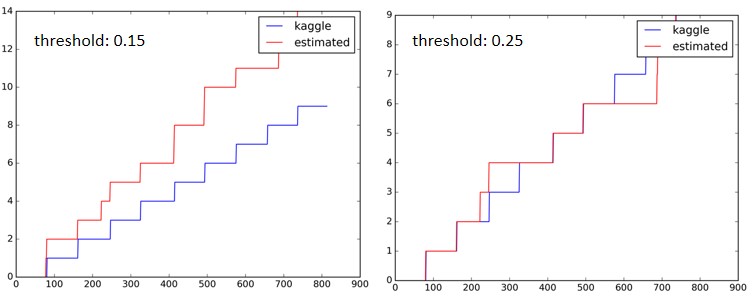}\label{s2}}
\caption{Temporal segmentation. (a): Blue lines represent the left and right wrist and elbow coordinates from the Convolutional Pose Machine output. Red lines are the Group Fused LASSO estimations. (b): switch estimations for two different threshold values, see the legends. Blue lines: the ground truth episode labels from 0 to 8 according to the Kaggle database. Red lines: the episode labels computed by the Group Fused LASSO.}\label{fig:switches}
\end{figure}

We have applied the Convolutional Pose Machine for each original frame independently to extract arm joint coordinates. The network was pre-trained mostly for frontal 2D body samples, yet it still shows reasonable performance for side views and for upper body configurations. As network outputs were extremely noisy across time, we tried to extract the underlying temporal segments with constant pose using the Group Fused LASSO (Sect.~\ref{ss:group_lasso}). Pose outputs were normalized and weighted by their scores. As CVXPY \citep{diamond2016cvxpy} did not yield a truly sparse solution, change point strengths were further thresholded. The result -- a nearly perfect match with ground truth labels, despite the large portion of noise -- for one of the videos is shown in Fig.~\ref{fig:switches}. 
Similar results can be achieved on the other videos. In all cases, the temporally segmented frames form \emph{groups} that we shall discuss later. 

\subsection{Episodic segmentation with Optical Flow} 

\begin{figure}[ht!]
\centering
\includegraphics[width=90mm]{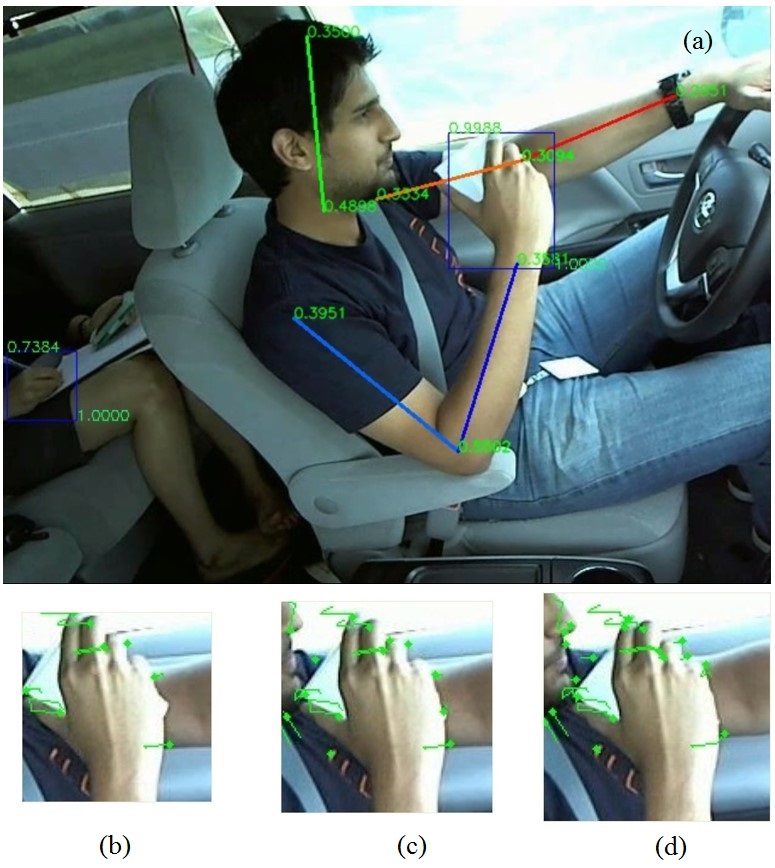}
\caption{Region-based Fully Convolutional Network hand-detector and Lucas\,--\,Kanade Optical Flow. (a): right hand (blue box), left and right arm poses (red and blue lines, respectively), head pose (green line), (b)-(d): three bounding boxes selected by our hand detector from subsequent frames. Green lines represent the motion vectors of the individual features.}\label{fig:OF}
\end{figure}

We applied Optical Flow (Sect.~\ref{ss:OF}) with the Region-based Fully Convolutional Network hand-detector (Sect.~\ref{ss:dns}) on raw videos. Fig.~\ref{fig:OF} illustrates the results and our point.
Subfigure (a) represents a close-to periodic motion within a larger image. It is labeled `drinking' in the Kaggle database: the driver is holding a cup, while he starts and stops drinking in an alternating manner. As described in Sect.~\ref{ss:OF}, the relevant portions of the images are segmented by our hand detector, the coherence of motion is made visible in subfigures (b)-(d) by the green arrows that show the motion of the individual feature points. The arrows change direction in the two drinking phases (not shown) and the Optical Flow itself can serve episodic segmentation, especially if Dense Optical Flow algorithms are used. Furthermore, similarly to the Group Fused LASSO example above, the bounding boxes connected by the optical flow form \emph{groups} in a natural fashion, to be detailed in Sect.~\ref{s:disc}.

\subsection{Rule-based combination of Object Detection and the Convolutional Pose Machine}\label{ss:correction}
\begin{figure}[ht!]
\centering
\subfigure[]{\includegraphics[width=30mm]{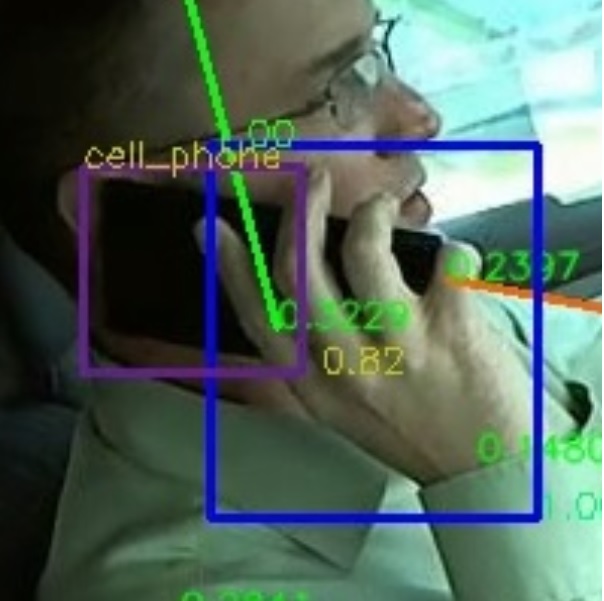}\label{qa}}
\subfigure[]{\includegraphics[width=30mm]{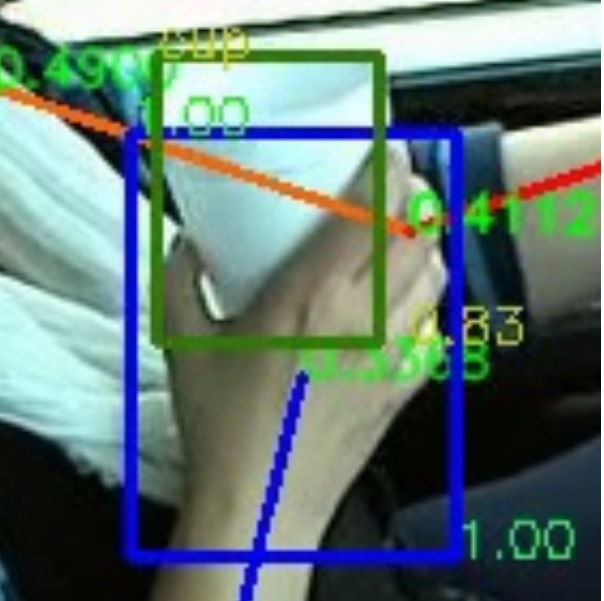}\label{qb}}
\subfigure[]{\includegraphics[width=30mm]{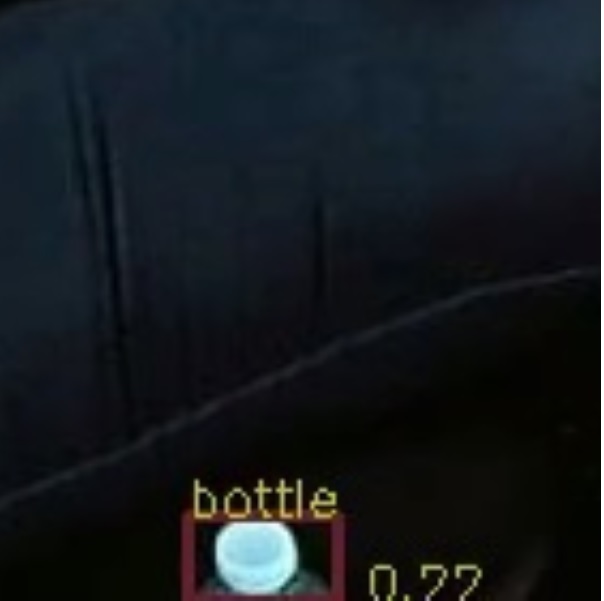}\label{qc}}\\
\subfigure[]{\includegraphics[width=90mm]{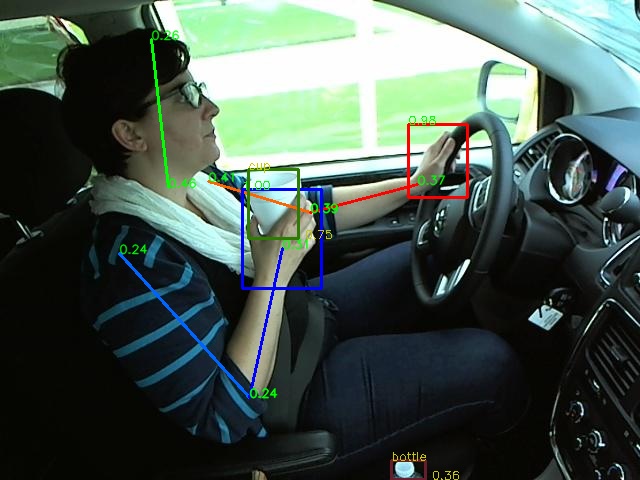}\label{qd}}
\caption{Object detection and the Convolutional Pose Machine in the Kaggle benchmark. (a): cellphone, (b): cup, (c): bottle, (d): the original image; the red and blue boxes and lines show the hand detector and the arm pose results (left and right, respectively), green line shows the head pose results.}
\label{fig:objects}
\end{figure}

\begin{figure}
\centering
\subfigure[]{\includegraphics[height=33mm]{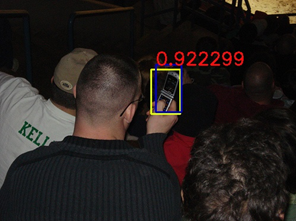}\label{a0}}
\subfigure[]{\includegraphics[height=33mm]{figures/hands2.png}\label{b0}}
\caption{Object detection samples with scores for cellphone. Both samples are from the MS COCO database.}
\label{fig:hands}
\end{figure}

The episodic description can be detailed if facial information predictors or object detectors are included (Sect.~\ref{ss:dns}), as shown in Fig.~\ref{fig:talking} and Fig.~\ref{fig:objects}, respectively.
Object detection is relatively weak for the Kaggle benchmark, as occluded objects are frequent, and there is a lack of rich datasets that would have such scenarios, e.g., objects in hand. Consider Fig.~\ref{fig:hands}, showing test results for the Faster R-CNN network from MS COCO (Sect.~\ref{ss:dns}).
In Fig.~\ref{a0} it is easy to recognize the cellphone. However, the cellphone can be inferred only from higher order knowledge in Fig.~\ref{b0} and the classification can be questioned. Both images are from the test dataset and the higher order knowledge is implicit in the labeling of the training database. 

We proceeded as follows. We `know' that labels are about human activities that are manifested by gaze that may either search or be focused on something, hands that provide information about the actual manipulation, and speech that we don't have in this benchmark. We used  Mittal’s Hand Dataset for hand detection and the VIVA Hand Detection Dataset for left and right hand classification. We employed the Region-based Fully Convolutional Network for the localization of the hands and a vanilla Convolutional Network for classifying if a hand is left or right. Results were encouraging, but left and right hands were frequently misclassified. Then we combined the outputs with those of the Convolutional Pose Machine (Sect.~\ref{ss:dns}). This is the direction of self-training; we/it can exploit inference methods based on our knowledge or knowledge bases that point to the strengths of classical artificial intelligence procedures.

Within the Kaggle database, there is a special scenario called `Safe Driving'. In this case, both hands are on the steering wheel and this is a limited spatial region. Furthermore, during all driving scenarios, one hand was always on that wheel; only very few frames are exceptions. The interplay between the Convolutional Pose Machine and the hand detectors together with backing knowledge forms the Cognitive Deep Machine as follows:
\begin{enumerate}
    \item the Convolutional Pose Machine gives high scores;
    \item the hand detector detects two or more hands with bounding boxes and the detectors have high scores;
    \item the node belonging to one of the wrists of the pose machine is close to one of the edges of the bounding box of the hand detector;
    \item the vector that starts at the node of the elbow and ends on the corresponding wrist points towards the center of one of the hand detectors;
    \item both hands are in the region of the wheel.
\end{enumerate}

In case of doubt or if higher certainty is needed, then one can require the following:
\begin{enumerate}
\setcounter{enumi}{5}
     \item the scores of the hand detector should be high;
     \item the distance between the estimated point of the wrist and the bounding box of the hand should be small.
\end{enumerate}
Further rigor could be introduced by means of the Optical Flow; one may require that the above conditions are fulfilled for a longer time interval, depending on the frame rate. Now, due to the high score of the Convolutional Pose Machine we know which hand is which, i.e., left or right. The variety of the set is limited and we can learn to recognize the left and the right hands \emph{on the steering wheel}. One may proceed from here, since for other scenarios one of the hands is on the steering wheel. In turn, backing knowledge says that the other hand, which is not on the steering wheel should be the other hand, provided that the pose machine supports the proposition. We note that there are two more people in the car; one is next to the driver and is working with camera. The other person is sitting on the back seat and is taking notes, or making phone calls, or may be texting. 
 
In the procedure we used one label set from the benchmark. Note, however, that this information can be dropped if we can recognize the steering wheel. The same procedure would follow. ImageNet has over 2,000 samples for this class\footnote{\url{http://image-net.org/synset?wnid=n04313503}}, so this training should be feasible.

\begin{figure}[ht!]
\centering
\subfigure[]{\includegraphics[width=120mm]{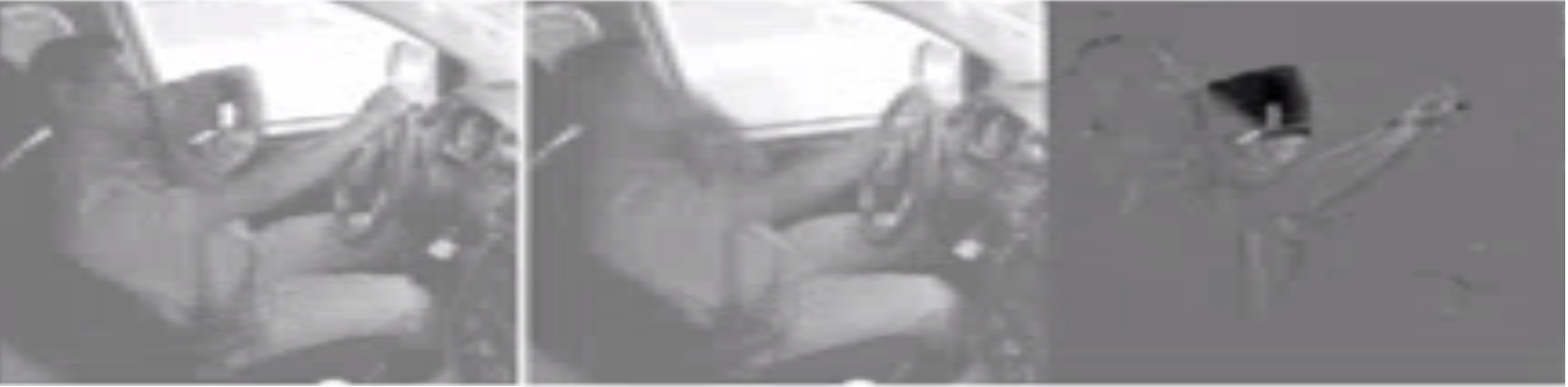}\label{va}}\\
\subfigure[]{\includegraphics[width=60mm]{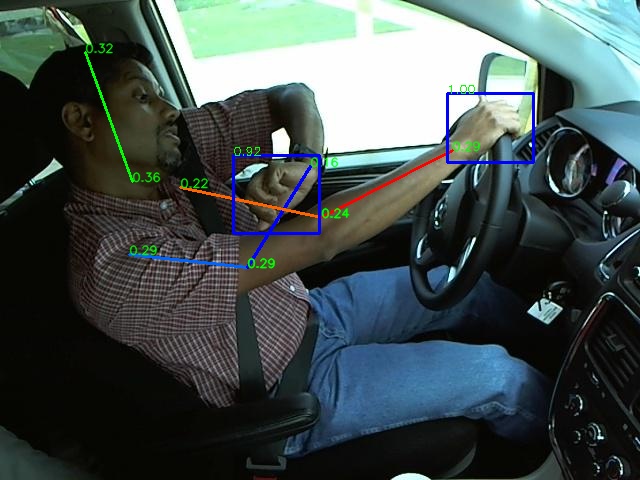}\label{vb}}
\subfigure[]{\includegraphics[width=60mm]{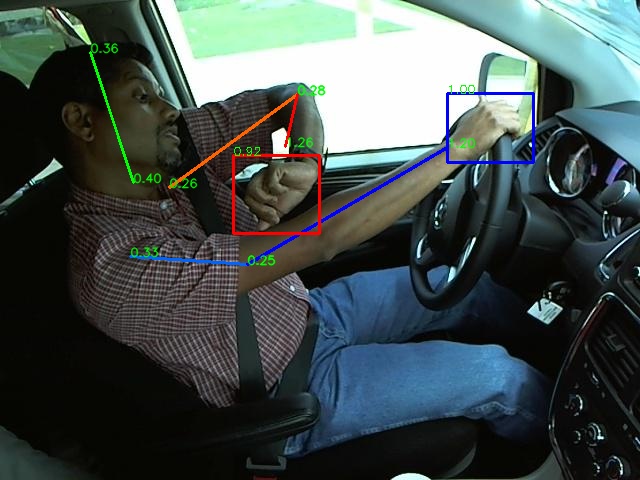}\label{vc}}
\caption{Warning from Robust Principal Component Analysis, inference based correction of the Hand Detector and the result of injection to the Convolutional Pose Machine. (a): the pose is very special, the full arm differs considerably from the average grey color. Left: image at low resolution, middle: the low dimensional projection, right: `outliers'. (b): Convolutional Pose Machine breaks down for both arms and both hands are classified as right hands. (c): one hand is on the wheel, its classification is trained and thus trusted in the scenario, the label of other hand is modified to left and these pieces of information are injected into the Convolutional Pose Machine giving rise to an improved pose estimation. This way, a training example is also generated for the Convolutional Pose Machine.}
\label{fig:corrected}
\end{figure}

The knowledge based method (one person has two hands and one of them is on the steering wheel) brings about good results for the Kaggle scenario. The information about the hands can be injected into the Convolutional Pose Machine that iterates heat maps about the suspected positions. The heat maps can be modified and the left and right hand positions of high probabilities can be injected that constrains the iterative procedure of the machine. We show some results in Fig.~\ref{fig:corrected}.

Finally, \emph{self-training} can be continued: given the backing knowledge and the required high scores:
\begin{enumerate}
\setcounter{enumi}{7}
     \item hand recognition is improved within the Kaggle benchmark; 
     \item a training example is collected for further tuning of the Convolutional Pose Machine within the scenarios of the benchmark.
\end{enumerate}

\section{Discussion}\label{s:disc}

\subsection{Self-training by deep networks having backing knowledge bases}

Outputs of the different input-output systems (e.g., deep networks) can be used for self-training. Our example is the Convolutional Pose Machine and the hand detector: we improved their performances by joining them (Sect.~\ref{ss:correction}). The Cognitive Machine can use high scoring poses and hand detector outputs for improving the recognition of the left and the right hands within the Kaggle scenarios. It may know that in the context of driving one hand is on the steering wheel in most of the time. It can use the information for giving proper labels to left and right hands and training itself. It can go further by using the same inference and inject the information about the left and/or the right hand  into the Convolutional Pose Machine. The derived pose can serve further training for the Convolutional Pose Machine within the benchmark scenarios. 

We note that the line of thoughts presented above is one option out of many. Other methods may serve the goals better, the goal being to have a joint agreement between the knowledge that concerns the actual context, the segmentation procedures provided by Group Fused LASSO, the warning signs of the RPCA method about outliers and the deep networks trained on other, more general databases. The applied heuristics, although proved to be successful, it is still accidental. Nonetheless, it fits reinforcement learning as we discuss in Sect.~\ref{ss:rl} that we discuss below.

\subsection{The engineering view: Back to rule based systems}

The machinery that we used simplified the Kaggle benchmark considerably. We can recognize the pose, separate the movements of the left and the right hands, one can track the face and the hand, too with high precision. Object detection can be raised to high levels, and thus the recognition of cell phones, cups/mugs should not be a big problem.  

The outcome of these considerations is that the classification according to the Kaggle labels is a simple ---~rule-based~--- task that does not require any further training. The meaning of a label is sufficient for proper classification. Furthermore, one can filter the ambiguous cases, e.g., when talking to the passenger and texting occur simultaneously. This is main reason why we think that the \emph{\citet{AI2030}} may overestimate the complexity of developing new applications. It is not deep technology alone that can serve new applications, but also the knowledge about the components and their relations, i.e., the semantic and ontological knowledge that have been collected over many years in history.

One question that emerges is the following: assuming that the labels are codes and thus have no meaning, could we tell what is happening? Temporal grouping by means of Group Fused LASSO and Optical Flow can help us. We have examples when the driver picks up the telephone, looks at the phone, manipulates the phone, lifts the phone to his ear and then talks. The interpretation of the scenario can be guessed by using ConceptNet\footnote{\url{http://conceptnet5.media.mit.edu/web/c/en/telephone}}. In turn, we can derive a number of rules from the Kaggle scenarios. One particular example is that dialing and holding the phone involves the same hand in driving scenarios.

Component searches and learning, followed by inferences can take us back to traditional AI. They enable fast learning, since one can learn new categories and relations through hearsay. Zebra is related to our horse, cow, dog illustration (Sect.~\ref{s:intro}). A giraffe is similar, it has spots and not stripes and its neck is much longer.  One can \emph{learn from the Kaggle labels}, too; they also provide information. For example, in ordinary communication, the separation of class `safe driving' may involve that the other categories are not safe. Similarly, we may suspect that there are phones, which are capable for text messaging, being common sense today, but not so about two decades ago. Such reasoning capabilities make decision making and, in turn, applications relatively easy to develop. If components are known then reasoning based upon them is sufficient in many cases. After all, what has been found by mankind over thousands of years, can be passed to a child in twenty years or so. The key, in our opinion, is in the knowledge of the components and deep neural networks can be trained   to recognize them. 

\subsection{The Gestalt view}

The principles that we applied are the hundred year old Gestalt Principles \citep{todorovic2008gestalt}, also called Gestalt Laws of Grouping, such as Proximity, Similarity, Continuity, Closure, and Connectedness. They can serve AI to discover spatio-temporal phenomena, or episodes, characterize and predict them, and if possible, compress them to meaningful concepts for human intelligence. Tacitly, we exploited determinism, when we used high scores and temporal relationships from Optical Flow and from Group Fused LASSO. The key hypothesis is that outlier detection and determinism \emph{together} are powerful tools for learning if we can recognize components and have (episodic) knowledge about their spatio-temporal relationships.

From the point of view of learning, if recognition has high score at a time, then it can help us to track the object and to learn its interactions, e.g., that a phone can be picked up. The more details we know, the less likely is that we make mistakes. This is also the Gestalt view.

\subsection{Cognition in the Cognitive Deep Machine}\label{ss:cognition}
The word cognition is used in many ways, including information processing, or acquiring knowledge, among others. In our view, a deep network is not capable of cognizing. We think that cognition is more than input-output mapping, it includes the capability of reasoning and utilizes reasoning for acquiring knowledge. Note that we are not using the word \emph{understanding}\footnote{\url{https://en.oxforddictionaries.com/definition/cognition}}, since it is beyond our present formulation.

Reasoning concerns considerations about the components, the reliability of the related observations as well as the spatio-temporal context. Cognition has at least five ingredients: (i) holistic and componentwise observations, e.g., by means of deep networks, (ii) outlier detection concerning those observations, (iii) reasoning about the best labels by means of the observed components, (iv) reasoning about the outliers and collecting such examples, and (v) self-training by means of the collected anomalies.

We went through such steps in the result section (Sect.~\ref{s:results}) and that can highlight the milestones of our approach. We note that up to this point we provided no algorithmic method, but only the name. This is so, since we believe that cognition is meaningless without goals. Thus, in our view, cognition is closely linked to reinforcement learning.

\subsection{Outlook: Reinforcement learning}\label{ss:rl}

Learning under supervision concerns input-output mapping. We think that there is a big difference between the supervised learning of components, like the case of the pose machine and the unsupervised searches for spatio-temporal structures. We think that the latter can be built stepwise from basic elements by means of Gestalt Principles, the example being the horse, the zebra, the giraffe, and the cow that look very similar, but differ in texture, the length of the neck, the shape of the body, including the mane and the hoof and the environment they live in (Sect.~\ref{s:intro}). The reduction of variables (here, components, or spatio-temporal structures) can put reinforcement learning into work, since reinforcement learning in the absence of such highly compressed factors restricted to those that might be relevant for decision making blows up exponentially. With such reduced spatio-temporal components, `factored reinforcement learning' becomes feasible as summarized by \citet{toser2015cyber} and the cited references therein. Important ingredient of the approach is the assumption on close to deterministic episodes that can be concurrent, can start and stop. Such concepts have been formulated within the  event learning framework for single episodic series \citet{szita2002varepsilon} and by \citet{szita2007learning} for multiple events.

Success stories on reinforcement learning using deep networks are already numerous, including Atari games \citep{mnih2015human} and the game Go \citep{silver2016mastering}. In turn, we conjecture that new applications built on deep learning technology based reinforcement learning, although technologically might be challenging regarding the sensory and the control systems, but may appear relatively fast as opposed to the view expressed in \emph{\citet{AI2030}}.

\section{Conclusion}\label{s:sum}

We considered the development of deep learning technology combined with traditional AI and component based reasoning. We provided examples through the State Farm Distracted Driver Detection benchmark. We argued that this benchmark requires components only and that they can be learned via deep learning in problems outside of the benchmark itself. Then, having these components, the benchmark requires no more learning, but reasoning, unsupervised anomaly detection, data collection related to the detected anomalies, searches for temporal segments, and finally self-training within the context of the Distracted Driver Detection problem. We conclude that novel applications can be tackled by AI and the bottleneck is in the sensory information and not in the learning system itself. We also argue that the component based construction has several advantages:
\begin{itemize}
    \item it can work if part of the information is missing, e.g., if the hand is not visible;
    \item it can be combined with reinforcement learning for the optimization of decision making;
    \item it can overcome the fragility of deep learning by means of component-wise reasoning;
    \item it can detect anomalies by means of inverting its own input-output system;
    \item it can look for spatio-temporal structures by means of the Gestalt Principles;
    \item it can self-train itself via its component based reasoning capabilities.
\end{itemize}
We justify our conclusion by highlighting that applications aim high level of determinism, training concerns a limited context, ontology and rule based systems are warranted, operation and potential errors can be modeled and run in virtual reality\footnote{\url{https://aimotive.com/what-we-do.jsp\#aikit}} and data can be collected in practice while searching for anomalies under less stringent conditions. The AI system can train itself step-by-step.  

\subsubsection*{Acknowledgments}

This work was partially supported by EIT Digital Grant No. 16527 on Cyber-Physical Systems for Smart Factories.


{\footnotesize 
\bibliographystyle{iclr2017_conference} 
\bibliography{ICLR}}

\end{document}